\documentclass[runningheads]{llncs}

\usepackage{hyperref}
\usepackage[acronyms]{glossaries}
\usepackage{mathtools}
\usepackage{tikz}
\usepackage{todonotes}
\usepackage{layouts}
\usepackage{algpseudocode}
\usepackage{algorithm}

\usetikzlibrary{external}
\usepackage{cite}
\usepackage{amsmath,amssymb,amsfonts}
\usepackage{graphicx}
\usepackage{textcomp}
\usepackage{xcolor}
\def\BibTeX{{\rm B\kern-.05em{\sc i\kern-.025em b}\kern-.08em
    T\kern-.1667em\lower.7ex\hbox{E}\kern-.125emX}}

\begin{document}

\title{On Restricting Real-Valued Genotypes\\ in Evolutionary Algorithms}

\author{J{\o}rgen Nordmoen\inst{1} \and
T{\o}nnes F. Nygaard\inst{1} \and
Eivind Samuelsen\inst{1} \and
Kyrre Glette\inst{1,2}}
\authorrunning{J. Nordmoen et al.}

\institute{Department of Informatics, University of Oslo, Norway\\ \email{jorgehn@ifi.uio.no} \and
RITMO, University of Oslo, Norway}

\newacronym{aco}{ACO}{Ant Colony Optimization}
\newacronym{ann}{ANN}{Artificial Neural Network}
\newacronym{cpg}{CPG}{Central Pattern Generator}
\newacronym{de}{DE}{Differential Evolution}
\newacronym{ea}{EA}{Evolutionary Algorithm}
\newacronym{ec}{EC}{Evolutionary Computation}
\newacronym{er}{ER}{Evolutionary Robotics}
\newacronym{map-elites}{MAP-Elites}{Multi-dimensional Archive of Phenotypic Elites}
\newacronym{moea}{MOEA}{Multi-Objective Evolutionary Algorithm}
\newacronym{nsga}{NSGA-II}{Non-dominated Sorting Genetic Algorithm-II}
\newacronym{pso}{PSO}{Particle Swarm Optimization}
\newacronym{qd}{QD}{Quality Diversity}

\maketitle

\begin{abstract}
    Real-valued genotypes together with the variation operators, mutation and crossover, constitute some of the fundamental building blocks of \acrlong{ea}s. Real-valued genotypes are utilized in a broad range of contexts, from weights in \acrlong{ann}s to parameters in robot control systems. Shared between most uses of real-valued genomes is the need for limiting the range of individual parameters to allowable bounds. In this paper we will illustrate the challenge of limiting the parameters of real-valued genomes and analyse the most promising method to properly limit these values. We utilize both empirical as well as benchmark examples to demonstrate the utility of the proposed method and through a literature review show how the insight of this paper could impact other research within the field. The proposed method requires minimal intervention from \acrlong{ea} practitioners and behaves well under repeated application of variation operators, leading to better theoretical properties as well as significant differences in well-known benchmarks.

\end{abstract}


\section{Introduction}
\Glspl{ea} are a class of optimization algorithms that take inspiration from nature~\cite{floreano2008bio}. By taking inspiration from biological concepts such as \textit{hereditary traits, genotype - phenotype distinction, mutation,} and \textit{survival of the fittest}, \glspl{ea} have been used to solve many challenging problems~\cite{chiong2012variants}. Many different encodings can be used to implement a genotype~\cite{Rothlauf2006}, among them is the real-valued genotype \textemdash a vector in $\mathbb{R}^n$~\cite{herrera1998tackling}. One often overlooked aspect of real-valued genotypes is the necessity to restrict the values to task specific bounds~\cite{wessing2013repair}. Restricting the values is fundamental since very few problems have infinite domains and because restrictions makes the search space feasible to explore~\cite{tanabe2019review}.

In this paper, we demonstrate the challenges of restricting real-valued genotypes and the effect these limitations have on the distribution of values in the genome. We start by introducing the theoretical problem and how restriction can affect the distribution. We then show that this problem is not simply a theoretical possibility; by optimizing benchmark functions we show that results diverge by solely varying the restriction function. We also look at the literature to gain an impression of how this problem could affect other \gls{ea} practitioners.

The contribution of our paper is a better understanding of how value restriction affects real-valued genotypes in \glspl{ea}. In addition we show that this is a topic deserving of more scrutiny by the wider \gls{ea} community through a literature overview of both well known \gls{ea} frameworks and a review of conference proceedings. By showing both the theoretical as well as the practical side of this challenge, we hope that other researchers become aware of the need for better restriction functions for real-valued genotypes and can utilize the results presented here in future research.

\begin{figure}[t]
    \centering
    \includegraphics{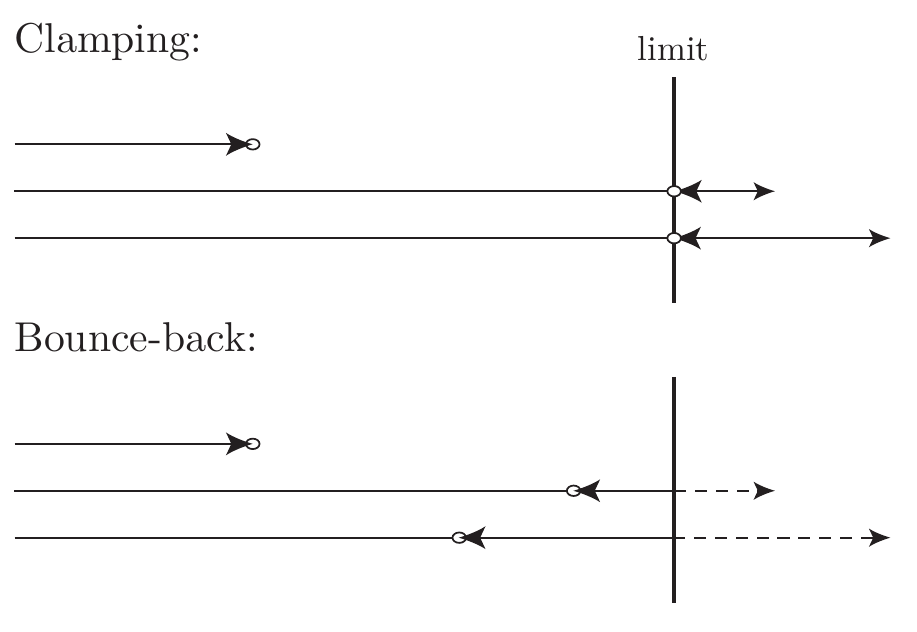}
    \vspace*{-15pt}
    \caption{By clamping the values in a real-valued genome the values will tend towards the limits of the bounds. The proposed \textit{Bounce-back} function reflects values that exceed the limits back inside the bounds, resulting in a uniform distribution inside the bounds.}
    \label{fig:illustration}
\end{figure}

\section{Background}
When designing a restriction function for real values, it is intuitive to create some form of a clamp method which ensures that a value does not exceed desired bounds\footnote{Like the `Clamp' method introduced in .NET Core: \url{https://docs.microsoft.com/en-us/dotnet/api/system.math.clamp?view=netcore-2.0} - accessed 18.05.2020}. In the literature few articles have been dedicated to understanding the effect a restriction function could have on the evolutionary process. In \cite{arabas2010experimental} the authors compare different restriction methods on the result of evolving with \gls{de} and showed that the results are impacted by the choice of restriction function. The paper \cite{wessing2013repair} extended on this work by understanding which type of inheritance model functioned with different restriction functions and as a by-product illustrated how these different restriction functions can alter the search landscape. A larger review of different components that contribute to the evolutionary process in \gls{de} were undertaken in \cite{tanabe2019review}, in this review article the authors showed that the choice of restriction function can have a significant impact on the results, a result backed by a very comprehensive comparison on different benchmark functions. However, even with these articles, that show significant impact on results solely on the basis of the restriction function, it is our impression that few \gls{ea} practitioners heed these warnings and make appropriate accommodations in regards to restricting real-valued genotypes.


To gain insight into the practises of restricting real-valued genotypes within the wider \gls{ea} community, we conducted a limited literature review. We first began with identifying some of the larger open-source frameworks for implementing \glspl{ea} to see how value restriction is implemented.

DEAP~\cite{DEAP_JMLR2012} is one of the more popular\footnote{Based on citations.} Python based frameworks for implementing \glspl{ea}. Looking through the source code of this framework, two functions implement genome value restriction, \textit{`cxSimulatedBinaryBounded'} and \textit{`mutPolynomialBounded'}. Both of these functions utilizes the \textit{Clamped} function to limit real-valued genomes. SFERESv2~\cite{Mouret2010} is another well known framework implemented in C++. It too implements value restriction for real-valued genotypes as can be seen in the genotype definition in \textit{`evo\_float.hpp'}. The restriction function is implemented in \textit{`put\_in\_range'} and is equivalent to the \textit{Clamped} function. Both of these frameworks cite the same source when it comes to their implementation of real-valued genomes and restriction, NSGA-II, a widely used \acrlong{moea}~\cite{deb2000fast}. Consulting the source-code\footnote{Source code available here: \url{http://www.iitk.ac.in/kangal/codes.shtml} - accessed 06.04.2020.} of this algorithm, it can be seen that this too implements restriction through the \textit{Clamped} function.

To evaluate whether or not the observations about evolutionary frameworks are representative for the community, we conducted a small review of the main proceedings from GECCO 2019 and the 2019 and 2020 EvoAPPS proceedings. To evaluate if the results of a paper could be susceptible to the challenges, identified in this paper, we first identified experiments utilizing real-valued genotypes\footnote{Note that papers utilizing \acrlong{pso} were excluded.}, we then tried to identify if the authors discuss their strategy for value restriction or if anything could be discerned from the source of the experiments. The summary of our results can be seen in Table~\ref{tab:literature_overview}.

\begin{table}
    \caption{Overview of real-valued genomes and limitation function in previous conference main proceedings.}
    \label{tab:literature_overview}
    \centering
    \begin{tabular}{|l|l|l|l|}
        \hline
        & GECCO'19 & EvoAPPS 2019 & EvoAPPS 2020 \\
        \hline
         Total number of papers & 173 & 42 & 44 \\
        \hline
         Uses real-valued genotype & 37 ($21\%$) & 11 ($26\%$) & 10 ($22\%$) \\
        \hline
         Comments on value restriction & 4 & 3 & 1 \\
        \hline
    \end{tabular}
\end{table}

From the overview, we can see that real-valued genotypes are used in a large fraction of papers in these previous conferences. However, of those that could be identified to use some form of restriction only four papers in GECCO'19, three papers in EvoAPPS 2019 and one paper from EvoAPPS 2020 were found. Of the four papers in GECCO'19 two used strategies that can mitigate the challenges identified in this paper~\cite{glasmachers2019challenges,nordmoen2019evolved}, with one of those being our contribution, while the other two papers used the \textit{Clamped} function. Of the three papers identified in EvoAPPS 2019 two would not be affected, one not using Gaussian mutation~\cite{iacca2019compact}, another is our previous contribution using the methods proposed here~\cite{nygaard2019evolving}, while the last one used the \textit{Clamped} method. From EvoAPPS 2020, one paper was identified to not be susceptible~\cite{pontes2020evodynamic} by only using uniform random mutation between allowable bounds instead of Gaussian mutation.

\section{Methods}
Several possible functions can be devised for limiting individual genes in real-valued genotypes~\cite{arabas2010experimental}. The easiest and most straight forward limitation function is to clamp the value to the given bounds. The function is defined as follows

\begin{equation}\label{eq:hard_limit}
    clamp(v, min, max) =
    \begin{cases}
    min \text{ if } v < min\\
    max \text{ if } v > max\\
    v \text{ otherwise}
    \end{cases}
\end{equation}
where $v$ is the value to limit and $[min, max]$ are the bounds to apply for the value. This function will be used as the baseline to compare against the proposed \textit{Bounce-back} method as it can be seen as the default when limiting real numbers\footnote{Method included in C++17: \url{https://en.cppreference.com/w/cpp/algorithm/clamp} - accessed 15.04.2020} and is the only one implemented in the \gls{ea} frameworks reviewed. We will refer to this function as \textit{Clamped} for the remainder of this text.

The proposed limit function, which we will call \textit{Bounce-back} restriction (also known as \textit{reflection}~\cite{arabas2010experimental}), is defined as follows
\begin{equation}\label{eq:soft_limit}
    \text{\it bounce-back}(v, min, max) =
    \begin{cases}
    min + (min - v) \text{ if } v < min\\
    max - (v - max) \text{ if } v > max\\
    v \text{ otherwise}
    \end{cases}
\end{equation}
where, again, $v$ is the value to limit and $[min, max]$ are the bounds to apply for the value. The effect of the \textit{Bounce-back} function is to redirect out-of-bounds values by the amount that the value is outside of the limits. The effect is illustrated in Figure~\ref{fig:illustration}. The function is independent of both mutation and crossover operators and can be applied in-between or after variation to ensure the genome is within given bounds. Compared to other restriction functions, such as \textit{wrapping}, \textit{re-initialization} and \textit{re-sampling}\cite{arabas2010experimental}, the \textit{Bounce-back} function results in values in the current vicinity of the solution while also having minimum computational impact.

One thing to note about the \textit{Bounce-back} function, as defined in Equation~\ref{eq:soft_limit}, is that it is not guaranteed to result in values within the given bounds. This can happen when the value $v > max + (max - min)$ or $v < min - (max - min)$. This can either be solved by continued application of the restriction function until the value is within bounds or limiting the difference in Equation~\ref{eq:soft_limit} with the \textit{Clamped} function. The later solution was utilized when limiting the distribution in Figure~\ref{fig:hist_limit}.

To demonstrate the unwanted properties of the \textit{clamp} function and how these properties are not present in the \textit{Bounce-back} function, we will first take an empirical approach to restricting real-valued genotypes with Gaussian perturbation.

\section{Empirical analysis}
\begin{figure}
    \centering
    \includegraphics{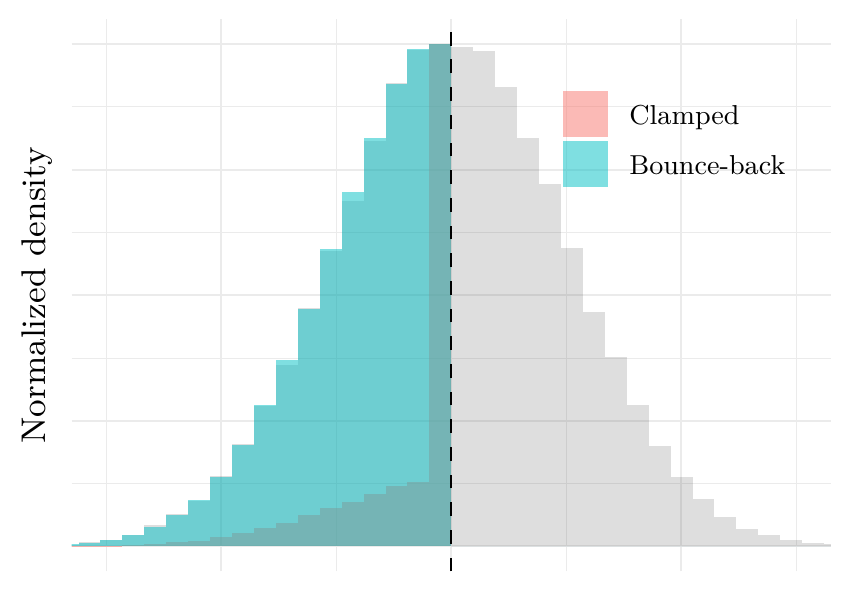}
    \vspace*{-10pt}
    \caption{The resulting distributions of applying the limit functions to a Gaussian distribution, in grey, with mean equal to the `limit' value marked with a dashed line. This is analogous to mutating a value at `limit' with Gaussian mutation and then restricting the value to be less than the `limit'. The \textit{Bounce-back} distribution follows the Gaussian distribution while the \textit{Clamped} distribution is skewed towards the `limit'.}
    \label{fig:hist_limit}
\end{figure}

To understand why using the \textit{clamp} function with real-valued genomes can be problematic we will begin by looking at how the function affects the Gaussian distribution. In Figure~\ref{fig:hist_limit} the result of applying the two limiting functions to the Gaussian distribution is shown. The grey area shows the original Gaussian distribution while the colored areas represent the respective limitation functions as applied. This example is akin to mutating a value that is on or near the bound with Gaussian perturbation before restricting the value to be less than the bound. As can be seen from the figure, the \textit{Bounce-back} function result in a distribution that is equivalent with the Gaussian distribution. Applying the clamping function results in a distribution that is heavily biased towards the `limit'. This shows the problem with simply clamping a value to the desired bounds, the value will be skewed towards the bounds because all values above the limit is restricted to become exactly the limit, as illustrated in Figure~\ref{fig:illustration}.

\begin{algorithm}
\begin{algorithmic}[1]
 \State \textbf{input} $n$
 \Comment{Number of values in genome}
 \State \textbf{input} $cycles$
 \Comment{Number of times to simulate mutation}
 \State \textbf{input} $\sigma$
 \Comment{Standard deviation of Gaussian perturbation}
 \State $genome\gets \mathcal{U}(0, 1)^n$
 \Comment{Create a vector of $\mathbb{R}^n$, uniformly distributed}
 \For{$n \leftarrow 0$ \textbf{to} cycles}
 \For{$i \leftarrow 0$ \textbf{to} $n$}
  \State $genome_i\gets \mathcal{N}(genome_i, \sigma)$
  \Comment{Apply Gaussian mutation}
  \If{$genome_i < 0.0$ \textbf{or} $genome_i > 1.0$}
   \State $genome_i\gets restrict(genome_i)$
  \Comment{Apply restriction, if outside bounds}
   \EndIf
  \EndFor
  \EndFor
 \State \textbf{return} $genome$ 
\end{algorithmic}
 \caption{Simulating mutation and value restriction of real-valued genotypes. `\textit{restrict}' is either the \textit{Clamped} or \textit{Bounce-back} function.}
 \label{alg:limit_gen}
\end{algorithm}

\begin{figure}
    \centering
    \includegraphics{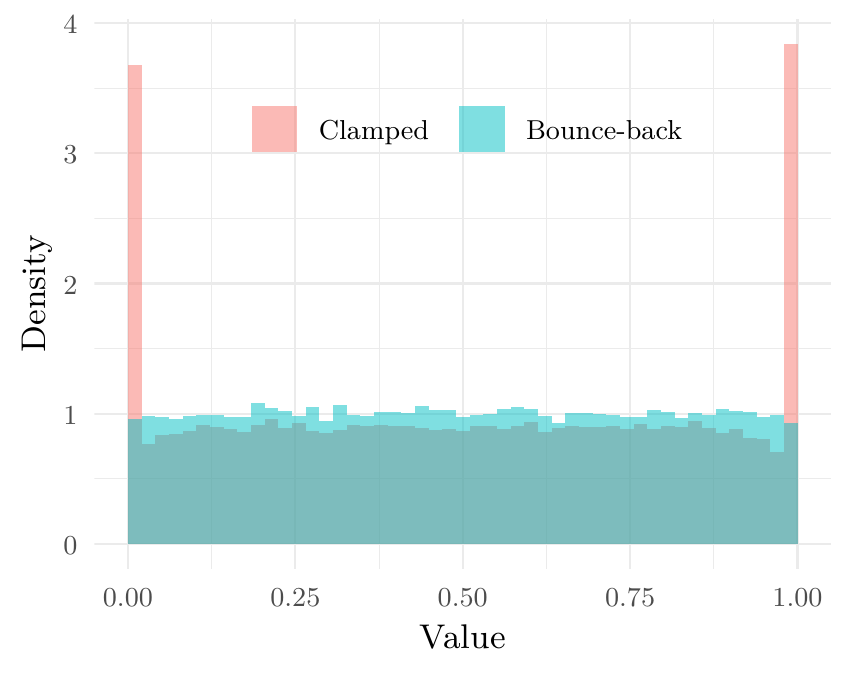}
    \vspace*{-10pt}
    \caption{The resulting distributions of generating $50000$ uniform values between $[0, 1]$, perturbing each value with a Gaussian, with $\sigma = 0.1$, and then limiting the value to the range $[0, 1]$. The perturbation-limitation cycle was run $100$ times.}
    \label{fig:hist_full}
\end{figure}

To visualize how the two limitation functions will affect a real-valued genome we created Algorithm~\ref{alg:limit_gen} which simulate how repeated application of Gaussian perturbation and restriction creates different distributions of values dependent on the limit functions. Figure~\ref{fig:hist_full} shows the distribution of values after running Algorithm~\ref{alg:limit_gen} with $n$ equal to $50 000$, $cycles$ set to $100$ and $\sigma = 0.1$. In the figure it can be seen that the distribution of the \textit{Clamped} function is heavily skewed towards the bounds while the \textit{Bounce-back} function has a much more uniform distribution.

To ensure that the distribution of the \textit{Bounce-back} limitation is uniformly distributed, we performed a One-sample Kolmogorov-Smirnov test~\cite{massey1951kolmogorov} comparing the two distributions in Figure~\ref{fig:hist_full} with a uniform distribution. The results show that the \textit{Clamped} distribution is statistically significant different from a uniform distribution, $p \ll 0.005$, while the \textit{Bounce-back} distribution is not significantly different, $p > 0.1$.

\begin{figure}
    \centering
    \includegraphics{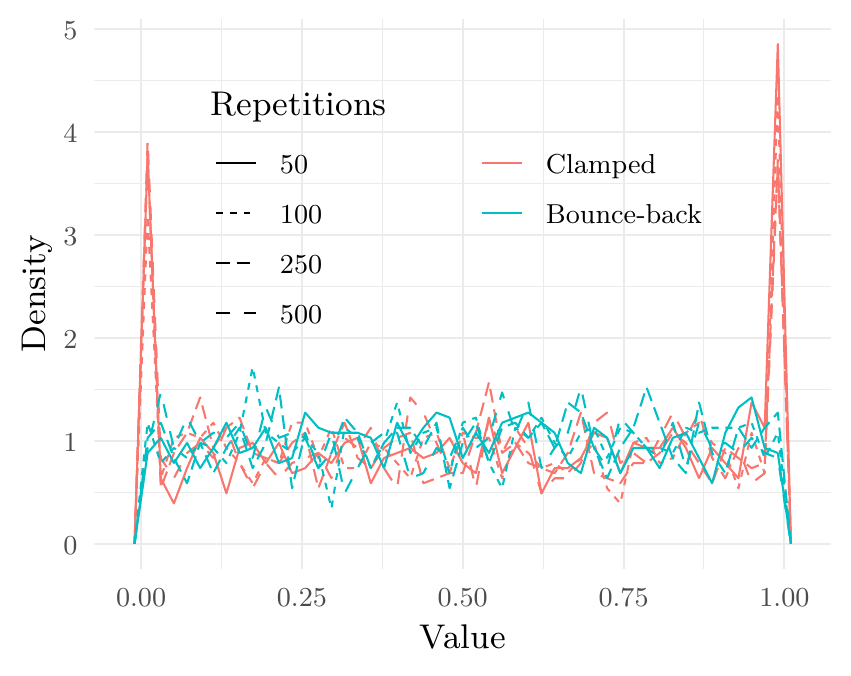}
    \vspace*{-10pt}
    \caption{A comparison of how the number of repetitions of the perturbation-limitation cycle affects the resulting distribution. The distribution in the figure were generated from $1000$ uniform random values and perturbed by a Gaussian with $\sigma = 0.1$.}
    \label{fig:comp_reps}
\end{figure}

\begin{figure}
    \centering
    \includegraphics{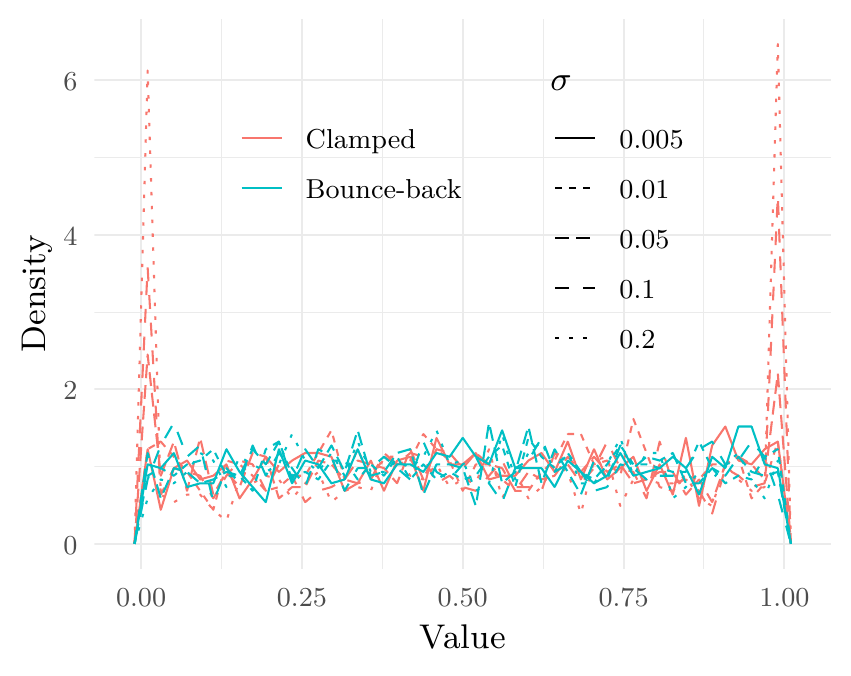}
    \vspace*{-10pt}
    \caption{A comparison of how the standard deviation, $\sigma$, of the Gaussian affects the resulting distribution. The distribution in the figure were generated from $1000$ uniform random values and perturbed $100$ times.}
    \label{fig:comp_sigma}
\end{figure}

To understand how the input parameters of Algorithm~\ref{alg:limit_gen} impact the output, we varied the two parameters, $cycles$ and $\sigma$. We will postulate that the number of reals, $n$, will not impact the underlying distribution and the effect of changing this parameter is to give a better or worse impression of the underlying distribution. The results of changing the input parameters are shown in Figure~\ref{fig:comp_reps} and Figure~\ref{fig:comp_sigma}, for number of $cycles$ and varying the standard deviation respectively. Changing the number of $cycles$, shown in Figure~\ref{fig:comp_reps}, does not have an effect on the resulting distribution which is as expected. Since the probability of the Gaussian mutation going out- or staying inside the bounds is symmetrical, the number of times mutation is applied should not effect the resulting distribution. For the standard deviation, shown in Figure~\ref{fig:comp_sigma}, the results are slightly different. Here we can see an effect of increasing $\sigma$, which can be explained as a larger part of the initial uniform distribution having a probability of going out of bounds. For the \textit{Clamped} function this results in more values restricted at the bounds as the standard deviation of the mutation increases. On the other hand, we can see that the \textit{Bounce-back} function is not affected by changes in mutation and continues to be uniformly distributed.

\section{Benchmark functions}
To illustrate the potential impact of the limitation function, we applied the two limitation functions to a selection of benchmark problems~\cite{yao1999evolutionary}. All functions were optimized with a single objective $(\mu + \lambda)$ \gls{ea} using tournament selection, Gaussian mutation, and no crossover operator. The benchmark problems are included to illustrate the challenge of limiting the genome to a specific range outside of theoretical considerations, as demonstrated in the previous section.

The following four functions are used as benchmark problems:
\begin{equation}\label{eq:griewank}
    \frac{1}{4000} \sum_{i = 1}^N x_i^2 - \prod_{i = 1}^N \cos\left(\frac{x_i}{\sqrt{i}}\right) + 1
\end{equation}

\begin{equation}\label{eq:rastrigin}
    10N + \sum_{i = 1}^N x_i^2 - 10 \cos\left(2 \pi x_i\right)
\end{equation}

\begin{equation}\label{eq:schaffer}
    \sum_{i = 1}^{N - 1} \left(x_i^2 + x_{i + 1}^2\right)^{0.25} * \left[\sin^2(50 * (x_i^2 + x_{i + 1}^2)^{0.10} + 1\right]
\end{equation}

\begin{equation}\label{eq:schwefel}
   418.9828872724339*N - \sum_{i = 1}^N x_i \sin\left(\sqrt{|x_i|}\right) 
\end{equation}

where $N$ is the size of the genome and $x_i$ is the value of gene $i$ in the genome. We will refer to equation~(\ref{eq:griewank}) as Griewank, equation~(\ref{eq:rastrigin}) as Rastrigin, equation~(\ref{eq:schaffer}) as Schaffer and equation~(\ref{eq:schwefel}) as Schwefel. The genotype was encoded as a vector of reals with a range of $[0, 1]$, before being transformed into the range required by each benchmark task. Genotypes are randomly initiated based on a uniform distribution. To generate data, we first optimized the parameters for each function, selecting the best mutation rate $\sigma$, tournament size and probability of applying Gaussian mutation to gene $i$ - $P(M | i)$. The final parameters used to generate data are shown in Table~\ref{tab:bench_params}.

\begin{table}
    \caption{Algorithm parameters for each benchmark function. Parameters marked with `\dag' were taken from \cite{brest2006self}.}
    \label{tab:bench_params}
    \centering
    \begin{tabular}{|l|l|l|l|l|}
         \hline
         & Griewank & Rastrigin & Schaffer & Schwefel \\
         \hline
         Generations\dag & \multicolumn{4}{c|}{2000} \\ 
         \hline
         $(\mu + \lambda)$\dag & \multicolumn{4}{c|}{$100 + 100$} \\
         \hline
         Size - $n$\dag & \multicolumn{4}{c|}{30} \\
         \hline
         Repetitions & \multicolumn{4}{c|}{50} \\
         \hline
         Tournament & \multicolumn{4}{c|}{10} \\
         \hline
         $P(M | i)$ & \multicolumn{4}{c|}{0.05} \\
         \hline
         Mutation - `$\sigma$' & 0.005 & 0.05 & 0.05 & 0.2 \\
         \hline
         Range & $\pm600$ & $\pm5.12$ & $\pm100$ & $\pm500$ \\
         \hline
    \end{tabular}
\end{table}

\begin{figure}
    \centering
    \includegraphics{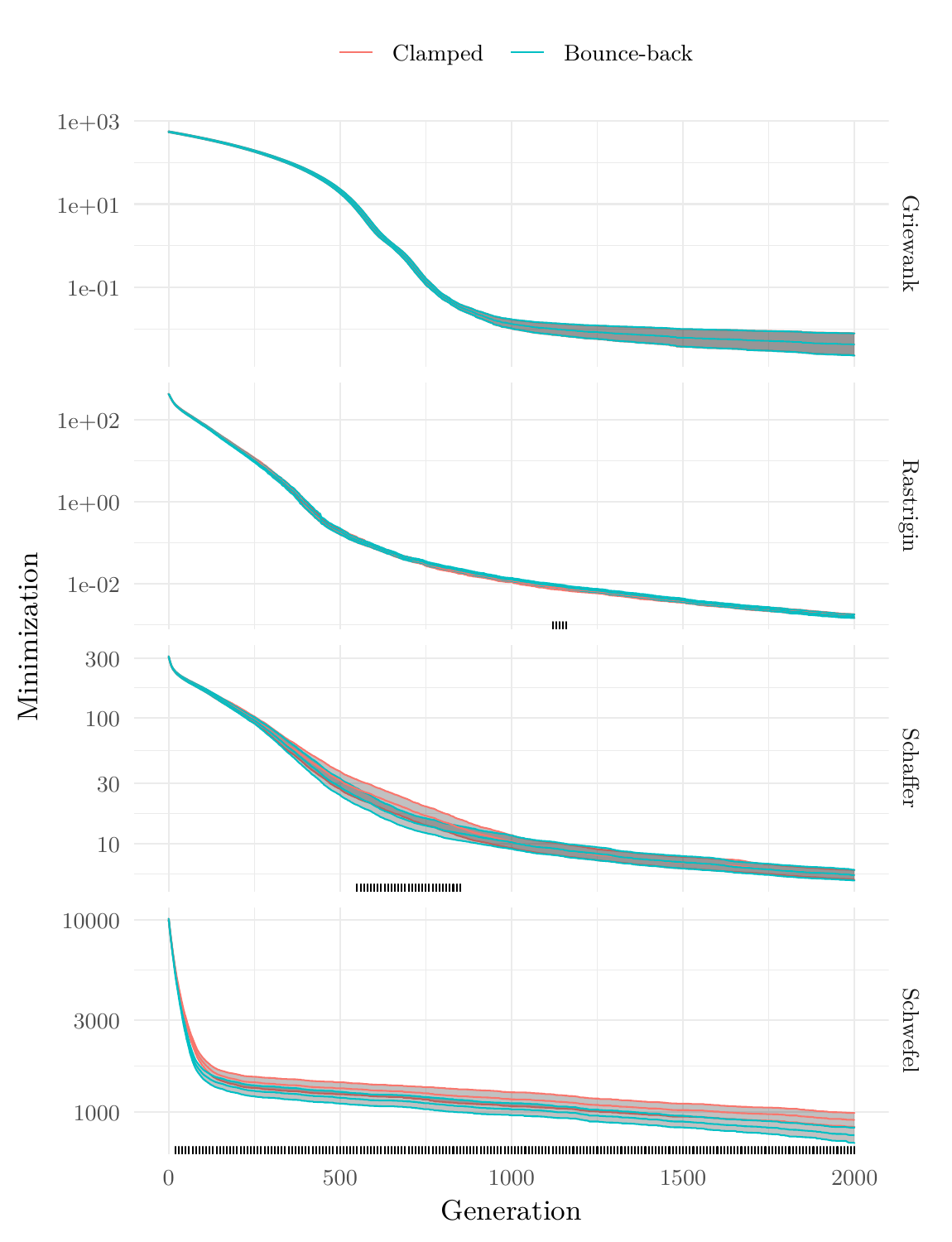}
    \vspace*{-10pt}
    \caption{The mean and $95\%$ confidence interval for the population minimum aggregated over $50$ runs of each benchmark function. Differences are marked at the bottom of each graph in black, the difference is statistical significant using a Wilcoxon Rank Sum test over an interval of $10$ generations adjusted for the number of tests using Holm correction.}
    \label{fig:bench_time}
\end{figure}

Figure~\ref{fig:bench_time} shows the mean and $95\%$ confidence interval of the population minimum for each benchmark function. Statistical significant differences between the two limitation functions are marked in black at the bottom of each plot and is the result of applying the Wilcoxon Rank Sum test on intervals of $10$ generations. To correct the p-value for the number of successive tests performed per row, Holm correction~\cite{holm1979simple} was applied. The result show that only one benchmark function is sufficiently affected to lead to different results, however, both Rastrigin and Schaffer show diverging results at some points before converging. Based on the individual search parameters, shown in Table~\ref{tab:bench_params}, we can also observe that larger $\sigma$ seems to induce larger differences.

\section{Discussion}
From the empirical analysis performed, it can be seen that the \textit{Clamped} function skews the distribution of values towards the bounds of the range, as shown in Figure~\ref{fig:hist_limit} and Figure~\ref{fig:hist_full}. This property can be challenging when the genotype is based on real values that are translated into task applicable ranges during the phenotype conversion. The proposed \textit{Bounce-back} function did not exhibit these properties resulting in no alteration of the underlying value distribution. When experimenting with different facets of mutation, Figure~\ref{fig:comp_reps} and Figure~\ref{fig:comp_sigma}, it was shown that the magnitude of the standard deviation had the most effect on the resulting value distribution. This can be understood as expanding the range of the Gaussian distribution applied at each value. As more values have the potential of being mutated near the bounds, more values will be restricted and end up at the extremities of the limit with the \textit{Clamped} function. This leads to the observation that using a larger standard deviation in the mutation operator require more careful thought to which restriction function to apply.

The results in Figure~\ref{fig:bench_time} shows that the choice of restriction function can have an effect on benchmark problems, and is not just a theoretical problem. The effect of the restriction function followed closely the magnitude of the standard deviation used for mutation, detailed in Table~\ref{tab:bench_params}, and illustrates the challenge that it can be difficult a priori to know the effect of the restriction function on any given task. One thing to point out about the benchmark tasks is that, except for \textit{Schwefel}, the target value for all genes is $x_i = 0$, which is in the middle of the range, away from the bounds. As the target value moves closer to the bounds, we would expect the effect of the restriction function to grow, which could also explain the larger difference observed with the \textit{Schwefel} benchmark function. With real-world tasks where the optimal value is difficult or impossible to confirm, it is even more challenging to a priori predict the effect and interactions of the restriction function, making it paramount that the restriction function has minimal impact on the optimization process.

The literature review conducted showed that two popular \gls{ea} frameworks utilizes the \textit{Clamped} function for restriction, and that while many papers published at GECCO'19 and EvoAPPS 2019 and 2020 utilized real-valued genotypes, few papers discussed their use of restriction functions. A deeper dive into the source of NSGA-II, a foundational algorithm within the \acrlong{moea} literature, also revealed the use of the \textit{Clamped} function which could affect re-implementations. Although limited, the literature review does underscore how the findings in this paper could impact the broader research field. As shown previously, the restriction function is not guaranteed to lead to significantly different results, however, it is difficult to predict the effect beforehand.

The results shown in this paper has focused on restricting real-valued genotypes in \glspl{ea}. One interesting direction to take this work in the future is to apply the same \textit{Bounce-back} technique on particles in \gls{pso}\cite{kennedy1995particle}. Because of the additional velocity attribute present in \gls{pso} the restriction function should take this into account when limiting so that the particles do not keep moving towards the bounds. In the same vein, it would be interesting to know if \gls{pso} is sensitive to the challenges presented in this paper, or if the collective behavior can mitigate the boundary effect shown in this paper.

\section{Conclusion}
In this paper we have shown that one of the most used functions for restricting real-valued genotypes behaves badly under repeated application of the variation operators. We therefore suggest a different function which result in a uniform distribution of values within the genome. The \textit{Bounce-back} function is shown both empirically and in practise to lead to better distributions and requires minimal intervention from existing \acrlong{ea} implementations, while having a minimal computational impact. We also conducted a limited literature review which illustrated that other practitioners in the field of \acrlong{ea}s could be susceptible to the complications detailed in this paper. We hope that by illuminating this problem other researchers will become aware of the requirements for restricting real-valued genotypes, and can hopefully mitigate it with minimal effort in future work.

\section*{Acknowledgement}
This work is partially supported by the Research Council of Norway through its Centres of Excellence scheme under grant agreements 240862 and 262762.

\bibliographystyle{splncs04}
\bibliography{limit}
\end{document}